\documentclass[letterpaper, 10 pt, conference]{ieeeconf}  % Comment this line out if you need a4paper

\IEEEoverridecommandlockouts                              % This command is only needed if 
\usepackage[usenames]{color}
\usepackage[svgnames]{xcolor}
\definecolor{DarkGreen}{rgb}{0,0.5,0}
\definecolor{DarkRed}{rgb}{0.75,0,0}

\usepackage{amsmath} % assumes amsmath package installed
\usepackage{nameref}
\usepackage{amssymb}
\usepackage{graphicx}
\usepackage{soul}
\usepackage{tikz}
\usetikzlibrary{shapes,snakes}
\usepackage{comment}
\usepackage{url}
\usepackage[ruled,vlined,linesnumbered]{algorithm2e}
\usepackage{balance}
\usepackage[font=small]{caption}

\usepackage{array}

\newcommand{\bee}{\begin{enumerate} \itemsep -1pt \topsep -2pt}
\newcommand{\eee}{\end{enumerate}}

\usepackage[flushmargin]{footmisc}

\usepackage[colorlinks]{hyperref}
\hypersetup{citecolor=Black}
\hypersetup{linkcolor=Black}
\hypersetup{urlcolor=DarkBlue}

\newcommand{\subparagraph}{}
\usepackage{titlesec}

\usepackage{caption}

\usepackage{mathtools}
\usepackage[capitalize]{cleveref}
\crefformat{equation}{(#2#1#3)}
\Crefformat{equation}{Equation~(#2#1#3)}
\Crefname{equation}{Equation}{Equations}

\usepackage{bm}
\usepackage{titlesec}

\definecolor{MydarkRed}{RGB}{183,21,33}
\definecolor{MydarkGreen}{RGB}{59,143,50}
\definecolor{Myred}{RGB}{255,0,0}
\definecolor{MyorangeDarker}{RGB}{255,127,42}
\definecolor{MygreenDark}{RGB}{55,200,55}
\definecolor{MyorangeDark}{RGB}{255,212,42}
\definecolor{MyblueLight}{RGB}{85,221,255}
\definecolor{Myblue}{RGB}{0,0,255}

\titlespacing{\section}{8pt}{7pt}{6pt}

\renewcommand{\baselinestretch}{.98}
\title{\LARGE \bf
Personalized Cancer Chemotherapy Schedule: a numerical comparison of performance and robustness in model-based and  model-free scheduling methodologies
}

\author{Jesus Tordesillas$^{1,*}$ and  Juncal Arbelaiz$^{2,*}$ \\  
	\thanks{$^{1}$J.~Tordesillas, MIT Department of Aeronautics and Astronautics \tt\ jtorde@mit.edu}
	\thanks{$^{2}$J.~Arbelaiz, MIT Department of Applied Mathematics. \tt\ arbelaiz@mit.edu}
	\thanks{$^*$ Both authors contributed equally to this work.}
}
\begin{document}
%\definecolor{orcidlogocol}{HTML}{A6CE39}

\maketitle
\thispagestyle{empty}
\pagestyle{empty}

%%%%%%%%%%%%%%%%%%%%%%%%%%%%%%%%%%%%%%%%%%%%%%%%%%%%%%%%%%%%%%%%%%%%%%%%%%%%%%%%
\begin{abstract}
Reinforcement learning algorithms are gaining popularity in fields in which optimal scheduling is important, and oncology is not an exception. The complex and uncertain dynamics of cancer limit the performance of traditional model-based scheduling strategies like Optimal Control. Motivated by the recent success of model-free Deep Reinforcement Learning (DRL) in challenging control tasks and in the design of medical treatments, we use \textit{Deep Q-Network} (DQN) and \textit{Deep Deterministic Policy Gradient} (DDPG) to design a personalized cancer chemotherapy schedule. We show that both of them succeed in the task and outperform the Optimal Control solution in the presence of uncertainty. Furthermore, we show that DDPG can exterminate cancer more efficiently than DQN presumably due to its continuous action space. Finally, we provide some insight regarding the amount of samples required for the training. 
\end{abstract}

%%%%%%%%%%%%%%%%%%%%%%%%%%%%%%%%%%%%%%%%%%%%%%%%%%%%%%%%%%%%%%%%%%%%%%%%%%%%%%%%

\section{INTRODUCTION and OBJECTIVES}

 Cancer is a common name that is given to a group of diseases that involve the repeated and uncontrolled division and spreading of abnormal cells. These abnormal tissues are called tumors \cite{cancerRL}. Early diagnosis and effective treatment improve the survival rate of these diseases. \\

The optimal treatment schedule and drug dose vary according to the stage of the tumor, the weight of the patient, the white blood cell levels (immune cells), concurrent illness and age of the patient. Thus, proper scheduling and personalization of the chemotherapy treatment are vital to reducing the mortality rate. This motivated the use of techniques originating in engineering fields -- such as optimal control -- to derive optimal drug dosing for cancer chemotherapy \cite{tumorModel}. A review of model-based scheduling strategies is provided in \cite{review}.\\

One of the main challenges associated with the study of cancer as a dynamical system is that it is known to be complex, nonlinear, and its mechanisms of action are uncertain. Consequently, first principle mathematical models may not be able to account for all the variations in the patient dynamics \cite{cancerRL}. \\

\begin{figure}[h!]
	\centering
	\includegraphics[width=\columnwidth]{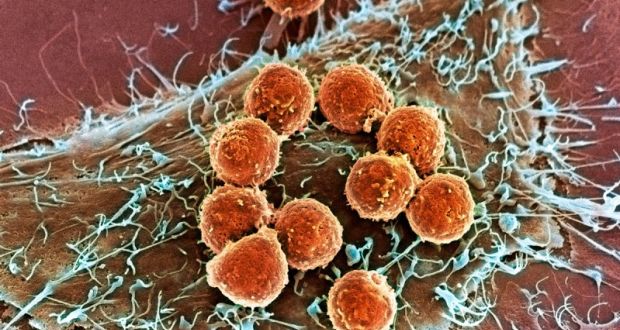}
	\caption{Immune cells (in orange) attached to a tumour cell (brown). This image was captured by a scanning electron microscope \cite{mskcc}. }
	%	\vskip -0.1in
	\label{fig:cells}
\end{figure} 

Motivated by the challenging nature of generating accurate models of cancer dynamics and the recent success stories of using RL for control \cite{science, control}, we use model-free Deep-Reinforcement Learning (DRL) algorithms to design a personalized cancer chemotherapy schedule. Particularly, we will use Deep Q-Network (DQN) \cite{DQN} (with a discrete action space) and Deep Deterministic Policy Gradient \cite{DDPG} (with continuous action space), and provide a comparison of their performance.\\

RL and DRL have been successfully applied in different medical treatment designs. For instance, in \cite{liu2017deep} DRL was used to determine the optimal treatment regimes from medical data.  \cite{tseng2017deep}  developed automated radiation adaptation protocols for Lung Cancer using DDPG. In \cite{cancerRL}, Q-learning was used to design a chemotherapy schedule with a discrete action space, and in \cite{Google} DL was used  for the detection of metastatic breast cancer. Moreover, \cite{engelhardt2019dynamic} used DDPG to control the drug dosing for suppressing cell growth modeled through a stochastic logistic growth model. \\

 The aim of this work is to develop \textit{in-silico} trials to compare the performance of model-based and model-free techniques in scheduling a personalized chemotherapy treatment. Particularly, our objectives include: \\
\begin{enumerate}
    \item Scheduling the personalized chemotherapy treatment by solving the optimal control problem to establish a baseline for comparison and evaluate whether DQN and DDPG provide similar performance.
     \item Gaining a qualitative understanding of the amount of training data (episodes) required by DQN and DDPG to perform similarly to the baseline policy. 
    \item Evaluating the robustness of the optimal controller, DQN and DDPG policies in the presence of different types of relevant uncertainties: parametric in the dynamics, parametric in the initial conditions of the model (diagnosis) and in the presence of a stochastic forcing term in the tumor cell population dynamics.
    \item Comparing the performance of DQN and DDPG in the scenarios described above.
\end{enumerate}

\section{MODEL}
In order to compute the optimal control policy and the reward in DRL, a mathematical model that captures the distribution and effects of the chemotherapy drug is required. A realistic model should address tumor growth, the reaction of the human immune system to the tumor growth, and the effects of chemotherapy treatment on immune cells, normal cells and tumor growth \cite{tumorModel}, \cite{review}.\\

We will simulate patient's response to the treatment through a pharmacological model of cancer chemotherapy, given by a nonlinear and coupled system of 4 deterministic Ordinary Differential Equations (ODEs) \cite{cancerRL}:\\

\begin{footnotesize}
\begin{equation}
\begin{aligned}
& \dot{N}(t)  = r_2 N(t)(1-b_2N(t))-c_4 N(t)T(t)-a_3N(t)C(t) \\
& \dot{T}(t)  = r_1 T(t)(1-b_1 T(t))-c_2 I(t) T(t) - c_3 T(t)N(t) -a_2 T(t)C(t)\\
& \dot{I}(t)  = s + \frac{\rho I(t) T(t)}{\alpha + T(t)}-c_1 I(t) T(t)- d_1 I(t)-a_1 I(t) C(t) \\
& \dot{C}(t)  = -d_2 C(t)+u(t)\\
& \text{with} \hspace{.1cm} N(0)= N_0, T(0)=T_0, I(0)=I_0, C(0)=C_0 \hspace{.1cm} \text{and} \hspace{.1cm} t \geq 0 
\end{aligned}
\end{equation}
\end{footnotesize}

\vspace{.2cm}

\noindent where $I(t)$ is the number of immune cells, $N(t)$ is the number of normal cells,   $T(t)$ is the number of tumor cells and $C(t)$ is the drug concentration. The \textit{action} or \textit{control} is the chemotherapy drug infusion rate $u(t)$ [$\text{mg} \cdot l^{-1} \cdot \text{day}^{-1}$], which we will schedule through Optimal Control and DRL, respectively. The initial conditions, $N_0, T_0, I_0 \hspace{.1cm} \text{and} \hspace{.1cm} C_0$ are determined according to the diagnosis. From now on, we will refer to this nonlinear model by $\mathbf{\dot{x}}=\mathbf{f}(\mathbf{x},t)$, where $\mathbf{\dot{x}} := \text{d}\mathbf{x}/ \text{d}t$ and $\mathbf{x}(t) := [N(t) \hspace{.15cm} T(t) \hspace{.15cm} I(t) \hspace{.15cm} C(t)]^{\text{T}}$. For the sake of visualization, an electron micrograph of immune and cancer cells is shown in Figure \ref{fig:cells}.\\

The model parameters and the corresponding values used in our simulations are provided in Table \ref{fig: params}. Note that the patient and his diagnosis determine both the initial conditions and the value of the model parameters. \\

\begin{table*}[h!]
  \centering
  	\includegraphics[width=1.6\columnwidth]{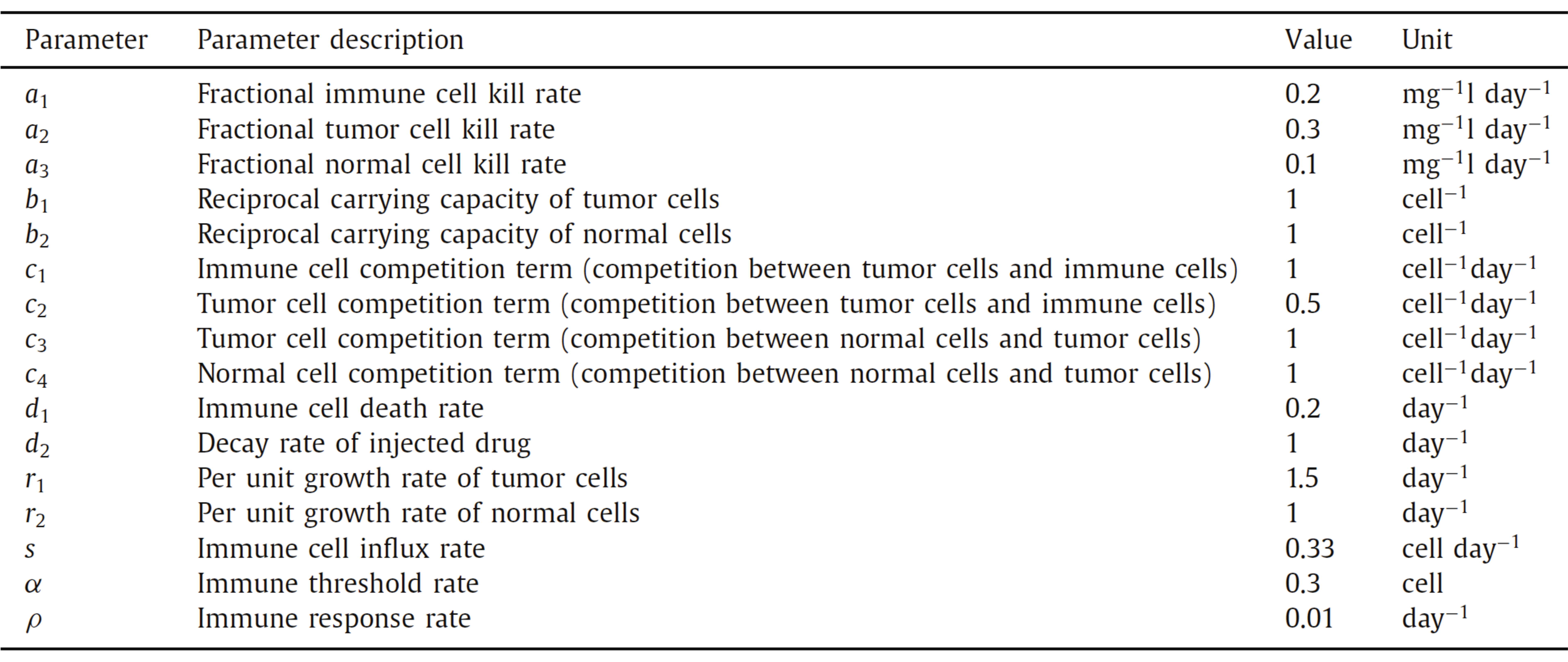}
  	\caption{Parameters of the model, description and the values used in our optimal control policy computation and training of DRL algorithms unless otherwise indicated. Extracted from \cite{cancerRL}.}
  	\label{fig: params}
\end{table*}

\iffalse
 \begin{figure*}[h!]
	\centering
	\includegraphics[width=1.6\columnwidth]{./table.pdf}
	\caption{Parameters of the model, description and the values used in our optimal control policy computation and training of DRL algorithms unless otherwise indicated. Extracted from \cite{cancerRL}. }
	\label{fig: params}
\end{figure*} 
\fi

As mentioned in the Introduction, in order to test the robustness of the optimal control and DRL policies, we simulated the previously presented system of ODEs with parametric uncertainty and also with a stochastic forcing term. The reader is referred to the next Section for further details on the uncertain scenarios considered.\\

\section{SCENARIOS CONSIDERED}

A personalized scheduling of chemotherapy treatment is fundamental to patient's recovery from the disease. When designing the treatment schedule, it is important to optimize the amount of drug used in order to regulate the potentially lethal side effects of chemotherapy, since often the patient's immune system weakens and becomes prone to life-threatening infections, diminishing its capability to eradicate cancer \cite{cancerRL}.\\

Consequently, we  obtained the Optimal Control and DRL policies  for two different cases: a preliminary and somewhat unrealistic \textit{Case 0}, in which the goal is to exterminate cancer regardless of the state of the rest of the cell populations. We used this case to validate our algorithms, codes and parameter and hypermarameter values. Then, we simulated a \textit{Case Patient}, in which cancer is eradicated while preserving a minimum population of normal and immune cells in order to guarantee patient's safety. In both cases, the initial condition was the same: $[1 \hspace{.2cm} 0.7\hspace{.2cm} 1  \hspace{.2cm} 0]^{\text{T}}$.\\

\subsection{Case 0}
The optimal control problem for \textit{Case 0} is formulated through the following nonlinear program:\\
\begin{equation}
\begin{aligned}
& \underset{u, t_f}{\text{min}}
  \int_0^{t_f} T(\tau) \hspace{.1cm} d\tau & &  \\
& \text{s. t.} & & \\
& \mathbf{\dot{x}} = \mathbf{f}(\mathbf{x},u)  & & \\
& \mathbf{x}(0) = \begin{bmatrix} 1 & 0.7 &  1 & 0 \end{bmatrix}^\text{T}& & \\
&  0\leq u \leq 10 \hspace{.1cm} \text{mg/(day l)} & & \\
& T(t_f) = 0
\end{aligned}
\label{eq:OC_case0}
\end{equation}
Both the policy $u(t)$ and the length of the treatment $t_f$ are determined by the optimal solution. The hard constraints correspond to the dynamics of cancer and the initial state of the patient, bounds in the drug-rate and the target of cancer eradication. The cost is the area enclosed by the tumor cell population (note that $T(t)$ is nonnegative) and the time axis, between $t=0$ and $t=t_f$. For the DRL algorithms, the reward used is $R=- dt\cdot T$, where $dt$ corresponds to the length of the timestep, and it is a fixed value. Note that $dt$ is included in the reward just to make the comparison with the cost functional used in the Optimal Control program straightforward.

\subsection{Case Patient}
In order to guarantee patient's safety during the treatment, additional state constraints are added to the program  (\ref{eq:OC_case0}). Particularly, the program is augmented with the constraints $N(t)\geq0.4$ and $I(t)\geq0.4$.  \\

In the case of the DRL algorithms, the constraints are imposed softly by modifying the reward function:

\begin{equation}
R=dt\cdot(- T - 0.5 \cdot [N<0.4] -0.5 \cdot [I<0.4])
\label{rewardConstrained}
\end{equation}

where $[\cdot]$ denotes the Iverson bracket. The last two terms of equation \eqref{rewardConstrained} penalize the violation of the constraints $N(t)\geq0.4$ and $I(t)\geq0.4$.  Note that, for the cases  in which the soft constraints are satisfied, this reward is the negative of the Riemann sum approximation of the area below the curve of the number of tumor cells $T$ (i.e. it is a rectangular approximation to the negative of the integral cost used in the optimal control problem). \\

Furthermore, in order to test the robustness of the obtained policies, different types of relevant sources of uncertainty will be considered regarding the diagnosis, growth-rate and dynamics of the tumor. The corresponding results are provided in Section \ref{sec:Results}.\\

\subsubsection{Parametric uncertainty: model parameter}

The per-unit growth-rate of the tumor, $r_1$, represents how aggressive the disease is. Thus, an accurate estimation of its value is important for the computation of the optimal control policy. We systematically varied its value and obtained the range for which the optimal control problem is feasible, as well as the sensitivity of the nominal optimal control policy to perturbations in the value of this parameter.\\

\subsubsection{Parametric uncertainty: initial condition} 

A wrong estimation of the initial size of the tumor $T_0$, will also have an impact on the performance of the optimal policy. We will systematically evaluate the robustness of the nominal policy to perturbations on $T_0$.\\

\subsubsection{Stochastic forcing in tumor dynamics }
We simulated a system of SDEs of the form: \\
 \begin{equation}
 d \mathbf{x_t}= \mathbf{f}(\mathbf{x_t}, t) dt+ \mathbf{G}(\mathbf{x_t},t) dW_t
 \end{equation}
 where the first term is the \textit{drift} and corresponds to the deterministic dynamics given by the ODEs, and the second one is the \textit{diffusion} term, modeled by a constant $\mathbf{G}$ in our case and applied only to the equation that gives the evolution of the population of tumor cells $T(t)$. $W_t$ denotes a Wiener process, common in the motion of cells. We simulated the SDEs using the \textit{Euler-Maruyama} scheme.

\section{METHODS AND ALGORITHMS}
\subsection{Optimal Control}

The general form of a finite-horizon optimal control problem in Bolza form is:
\begin{equation}
\begin{aligned}
& \underset{\mathbf{u} \in \mathcal{U},T}{\text{max}} \hspace{.2cm}
  J = \int_0^{T} F(\mathbf{x}(t),\mathbf{u}(t),t) \hspace{.1cm} dt + S(\mathbf{x}(T), T)& &  \\
& \text{s. t.} & & \\
& \mathbf{\dot{x}} = \mathbf{f}(\mathbf{x}(t),\mathbf{u}(t))  & & \\
& \mathbf{x}(0) = \mathbf{x}_0& & \\
&  \mathbf{g}(\mathbf{x}(t),\mathbf{u}(t),t) \geq \mathbf{0} & & \\
& \mathbf{h}(\mathbf{x}(t),t) \geq \mathbf{0} & & \\
& \mathbf{a}(\mathbf{x}(T), T) \geq \mathbf{0}& & \\
& \mathbf{b}(\mathbf{x}(T),T) = \mathbf{0}
\end{aligned}
\label{eq:standardOCproblem}
\end{equation}

 Inspection of the program (\ref{eq:OC_case0}) reveals that it constitutes a particular case of problem (\ref{eq:standardOCproblem}) and thus can be solved using standard optimal control techniques. Something to note is that the control solution provided by (\ref{eq:standardOCproblem}) will usually be open-loop (i.e. the control policy will be a function of time, as opposed to feedback or closed-loop control laws, in which the control is a function of the state. Usually this last case is more desirable since it accounts for discrepancies between the real dynamics and the predictions made by the model and corrects the control input accordingly, making the controller more robust to uncertainty).\\

\subsection{Deep Reinforcement Learning}
DDPG and DQN algorithms are considered. Both DDPG and DQN share two common features: they are \textit{off-policy} and \textit{model-free} algorithms. \textit{Off-policy} means that the behaviour policy used is not the same as the policy being improved. This allows the use of a memory replay, and the use of any exploration strategy. \textit{Model-free} means that the algorithm does not try to estimate the transition matrix of the dynamics of the environment $T(s_{t+1}|s_t,a_t)$. Instead, it estimates the optimal policy or value function directly.\\

The differences between these two algorithms are highlighted below. \\

\subsubsection{\underline{\textbf{Deep Q Network (DQN})}}

Deep Q Network was proposed in \cite{DQN}, and the main difference with respect to standard Q-learning is the use of a neural network to approximate the action-value function $Q(s,a)$. The DQN algorithm is shown in Alg. \ref{algo: algorithm_DQN} (taken from \cite{DQN}).\\

The inclusion of a neural network, usually leads to an unstable training due to two factors: the correlation between samples and the non-stationary targets. These two challenges are addressed by DQN using \cite{stanford_notes}:

\begin{itemize}
    \item \textbf{An experience replay}: In a replay buffer, a dataset of tuples $(s_i, a_i, r_i, s_{t+1})$ are saved. During the training, the agent will randomly sample mini-batch samples from this replay buffer (line \ref{use_of_D} of the algorithm). This allows a stabilization of the training process, and a better approximation to i.i.d samples removing the correlation between them. 
    \item \textbf{Fixed Q-targets}: During several updates, the target network weights used in the target calculation are fixed (lines \ref{reset_weights1}, \ref{reset_weights2}, \ref{reset_weights3}).\\
\end{itemize}

DQN is mainly used for problems with a discrete action space. However, some applications make use of the Normalized Advantage Function (NAF) to apply DQN in continuous action spaces \cite{gu2016continuous}. \\

\subsubsection{\underline{\textbf{Deep Deterministic Policy Gradient (DDPG)})}}

\begin{figure}[h!]
	\centering
	\includegraphics[width=0.7\columnwidth]{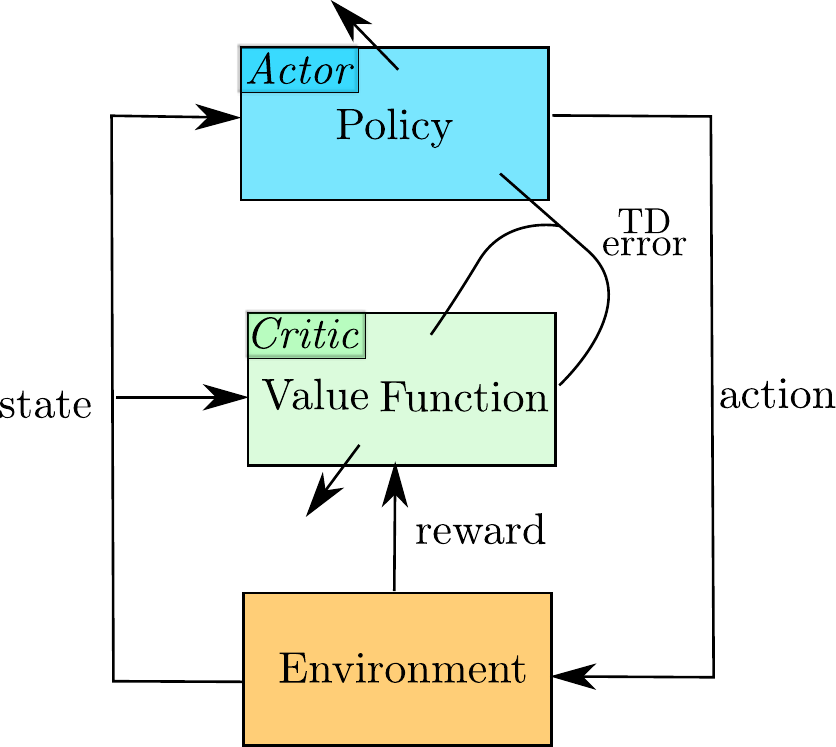}
	\caption{Actor and Critic used in DDPG. Adapted from \cite{sutton2018reinforcement}. }
	%	\vskip -0.1in
	\label{fig:ddpg}
\end{figure}

DDPG was proposed in \cite{DDPG}. The algorithm used by DDPG is shown in Alg. \ref{algo: algorithm_DDPG} (taken from \cite{DDPG}), and its main characteristics are these ones \cite{DDPG}, \cite{sutton2018reinforcement}, \cite{ddpg_tutorial}:

\begin{itemize}
\item \textbf{Policy-gradient}: DDPG tries to estimate the gradient of the expected return, and the policy is updated using this estimate. 
\item \textbf{Actor-critic}: DDPG has two different structures (see also Fig. \ref{fig:ddpg}) :
   \begin{itemize}
     \item Actor: The actor contains the policy function. It takes the state of the environment as input, and produces an action. 
     \item Critic: From the state of the environment and the reward received by the action taken by the actor, the critic produces the temporal difference error. This error is used to update both the actor (line \ref{actor}) and the critic (line  \ref{critic}). \\
   \end{itemize}
\end{itemize}
   
\noindent Both the actor and critic described above are represented using neural networks.

\begin{algorithm}[t]
	\footnotesize
	
	\DontPrintSemicolon
	
	\SetKwFunction{FMain}{\textcolor{ForestGreen}{\textbf{Replan}}}
	\SetKwProg{Pn}{Function}{:}{\KwRet}
	
	   Initialize replay memory D to capacity N\;
	   Initialize action-value function Q with random weights $\theta$\; \label{reset_weights1}
	   Initialize target action-value function $\hat{Q}$ with random weights\; \label{reset_weights2}
	   $\theta^{-}=\theta$ \;
	   \For{episode=$1:M$}{
	        Initialize sequence $s_1=\{x_1\}$ and preprocessed sequence $\phi_1=\phi(s_1)$\;
	        \For{t=$1:T$}{

	            With probability $\epsilon$ select a random action $a_t$\;
	            otherwise select $a _ { t } = \operatorname { argmax } _ { a } Q \left( \phi \left( s _ { t } \right) , a ; \theta \right)$\;
	            Execute action $a_t$ in emulator and observe reward $r_t$ and image $x_{t+1}$\;
	            Set $s_{t+1}=s_t,a_t,x_{t+1}$ and preprocess $\phi _ { t + 1 } = \phi \left( s _ { t + 1 } \right)$\;
	            Store transition $\left( \phi _ { t } , a _ { t } , r _ { t } , \phi _ { t + 1 } \right)$ in $D$\; \label{use_of_D}
	            Sample random minibatch of transitions $\left( \phi _ { j } , a _ { j } , r _ { j } , \phi _ { j + 1 } \right)$ from $D$\;
	            Set $y_{j}=\left\{ \begin{array}{cc}
r_{j}\text{ if episode terminates at step }j+1\\
r_{j}+\gamma\max_{a^{\prime}}\hat{Q}\left(\phi_{j+1},a^{\prime};\theta^{-}\right) & \text{ otherwise }
\end{array}\right.$\;            
	            
	            Perform a gradient descent step on $\left( y _ { j } - Q \left( \phi _ { j } , a _ { j } ; \theta \right) \right) ^ { 2 }$ with respect to the network parameters $\theta$\;
	            Every C steps reset $\hat { Q } = Q$ \label{reset_weights3}

	        }
	   }

\normalsize
	\caption{DQN \label{IR}}
	\label{algo: algorithm_DQN}
\end{algorithm}

\begin{algorithm}[t]
	\footnotesize
	
	\DontPrintSemicolon
%	\KwData{$A$, $G_{term}$, $\mathcal{O}$, $\mathcal{F}$, $\mathcal{U}$, $\delta_{a}<\delta_{b}<1 $, $\alpha_{0}>0$, $R_{b}>R_{a,max}>R_{a,min}>0$,   }
	%\KwResult{Next jerk-controleld primitive to execute}
	%	\textbf{Initialization} (only once): $JPS_2{old} \leftarrow JPS(A, G)$ //
	
	\SetKwFunction{FMain}{\textcolor{ForestGreen}{\textbf{Replan}}}
	\SetKwProg{Pn}{Function}{:}{\KwRet}
%	\Pn{\FMain{}}{
	
	   Initialize critic network $Q(s,a|\theta ^Q)$ and actor $\mu(s|\theta^\mu)$ with weights $\theta^Q$ and $\theta^{\mu}$.\;
	   Initialize target network $Q'$ and $\mu'$ with weights $\theta^{Q'}\leftarrow \theta^{Q}$, $\theta^{\mu'}\leftarrow \theta^{\mu}$\;
	   Initialize replay buffer $R$\;
	   	\For{episode=$1:M$}{ 
	   	Initialize a random process $\mathcal{N}$ for active exploration\;
	   	Receive initial observation state $s_1$\;
	   	\For{$t=1:T$}{
	   	Select action $a_t=\mu(s_t|\theta^{\mu})+\mathcal{N}_t$ according to the current policy and exploration noise\;
	   	Execute action $a_t$ and observe reward $r_t$ and new state $s_{t+1}$\;
	   	Store transition $(s_t, a_t, r_t, s_{t+1})$ in $R$\;
	   	Sample a random minibatch of $N$ transitions $(s_i, a_i, r_i, s_{i+1})$ from $R$\;
	   	Set $y_i=r_i+\gamma Q'(s_{i+1},\mu'(s_{i+1}|\theta^{\mu'})|\theta^{Q'})$\;
	   	Update critic by minimizing the loss: $L = \frac { 1 } { N } \sum _ { i } \left( y _ { i } - Q \left( s _ { i } , a _ { i } | \theta ^ { Q } \right) \right) ^ { 2 }$\; \label{critic}
	   	Update the actor policy using the sampled policy gradient:
	   	$\nabla_{\theta^{\mu}}J\approx\frac{1}{N}\sum_{i}\nabla_{a}Q\left.(s,a|\theta^{Q})\right|_{s=s_{i},a=\mu\left(s_{i}\right)}\nabla_{\theta^{\mu}}\mu\left.(s|\theta^{\mu})\right|_{s_{i}}$ \; \label{actor}
	   	   	Update the target networks:\; 
$\begin{array} { l } { \theta ^ { Q ^ { \prime } } \leftarrow \tau \theta ^ { Q } + ( 1 - \tau ) \theta ^ { Q ^ { \prime } } } \\ { \theta ^ { \mu ^ { \prime } } \leftarrow \tau \theta ^ { \mu } + ( 1 - \tau ) \theta ^ { \mu ^ { \prime } } } \end{array}$	
	   	} 

	}
%	}

	\normalsize
	\caption{DDPG \label{IR}}
	\label{algo: algorithm_DDPG}
\end{algorithm}

\section{IMPLEMENTATION}

\subsection{Optimal Control}

We solved the optimal control problem using direct collocation methods \cite{beets}, which transcribe the continuous dynamics and control functions to a finite set of algebraic variables and then solve a high-dimensional non-linear program (NLP). We used the MATLAB implementation of ICLOCS2 \cite{iclocs}, the open-source Imperial College London Optimal Control software (available for download \href{http://www.ee.ic.ac.uk/ICLOCS/#}{here}) in conjunction with the open-source NLP solver IPOPT \cite{ipopt}. We modified the open-source code and introduced the dynamics of cancer (the ODE system), cost functional and constrains, as well as the desired options and tolerances for the solvers.\\

We used h-methods for the transcription, particularly the Hermite-Simpson method with automatic mesh refinement and an initial number of 200 nodes. Regarding numerical tolerances, we allowed errors of up to $10^{-2}$ in the state and control bounds, and in the terminal condition of cancer eradication. The optimal control solution for both the \textit{Case 0} and \textit{Case Patient} is provided in Section \ref{sec:Results}. 

\textcolor{red}{}
\subsection{Reinforcement Learning}

To implement the DRL algorithms, we used the open-source library ChainerRL \cite{chainerRL}. Moreover, we created an environment using the OpenAI Gym \cite{OpenAI_gym} framework. This environment takes an action and the current state, and returns the next observation and the reward obtained. In this environment, we  implemented the system of ODEs, and solved it using the numerical integration methods provided by Scipy \cite{scipy}. \\

The values of the most relevant parameters of the DQN and DDPG implementations are shown in Table \ref{tab:parameters}. Note that we used $\gamma\approx 1$ to match the reward used in Optimal Control  as well as possible.\\

To improve the convergence and the training times, all the states, actions and the reward were normalized to values in $[0,1]$, and CuPy was used.\\

Note that both in DDPG and DQN, it is required to select the size of the time steps taken by the agent during an episode. We used a time step size of $0.3$ days, which achieved a reasonable training time. 

\begin{table}[]
\caption{Parameters used in DQN and DDPG} \label{tab:parameters}
\centering
\begin{tabular}{|l|l|l|}
\hline
\textbf{Parameter}     & \textbf{DQN}    & \textbf{DDPG} \\ \hline \hline
%Actor learning rate               & $10^{-4}$  \\ \hline
%Critic learning rate              & $10^{-3}$  \\ \hline
Replay Start Size      & $10^{3}$     & $5\cdot10^{3}$  \\ \hline
Layers                 & 2                 & 3  \\ \hline
Hidden Units per layer & 100               & 300   \\ \hline
Discount Factor $\gamma$                  & 0.99             & 0.995\\ \hline
%Mini-batch size        & 200               & 200  \\ \hline
\end{tabular}
\end{table}

%We also tried $\gamma=1$  but the algorithm did not converge. 

\section{RESULTS}
\label{sec:Results}

\subsection{Case 0}
\begin{figure}[h!]
	\centering
	\includegraphics[width=\columnwidth]{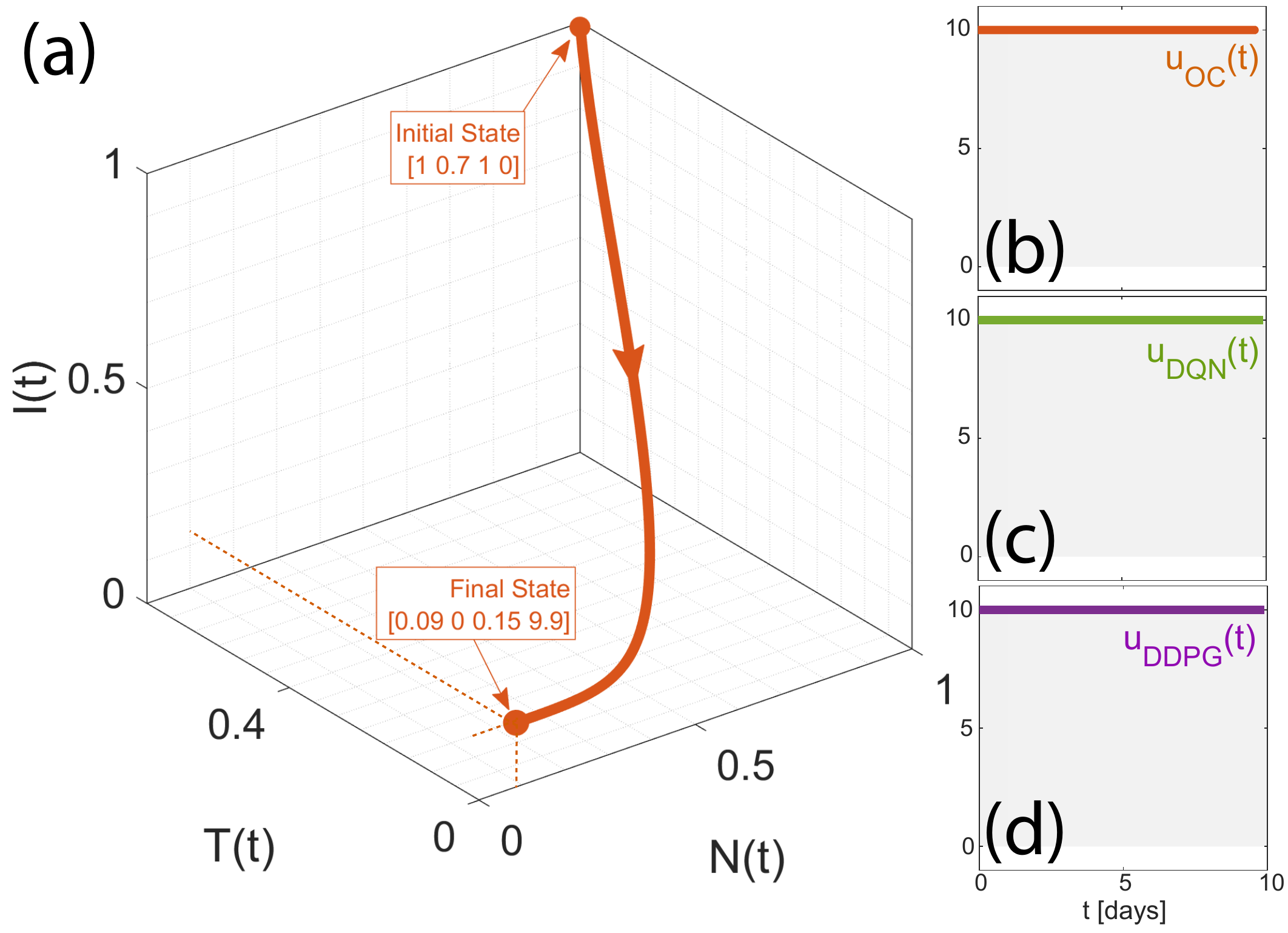}
	\caption{Results for the preliminary \textit{Case 0}. (a) Trajectories of the states when the optimal drug-rate is applied. Policies provided by (b) O.C. (c) DQN (d) DDPG. Note that (b)-(d) match. The shaded regions represent the feasible control region. }
	\label{fig:Case0_results}
\end{figure}

The results for the preliminary \textit{Case 0} (in which there are not constraints on $N(t)$ and $I(t)$) are shown in Fig. \ref{fig:Case0_results}. Note that the policy found by all the algorithms (O.C., DDPG and DQN) is the same: apply the maximum allowed drug infusion rate until the cancer is exterminated, which is the expected solution.
\begin{figure*}[h!]
	\centering
	\includegraphics[width=2\columnwidth]{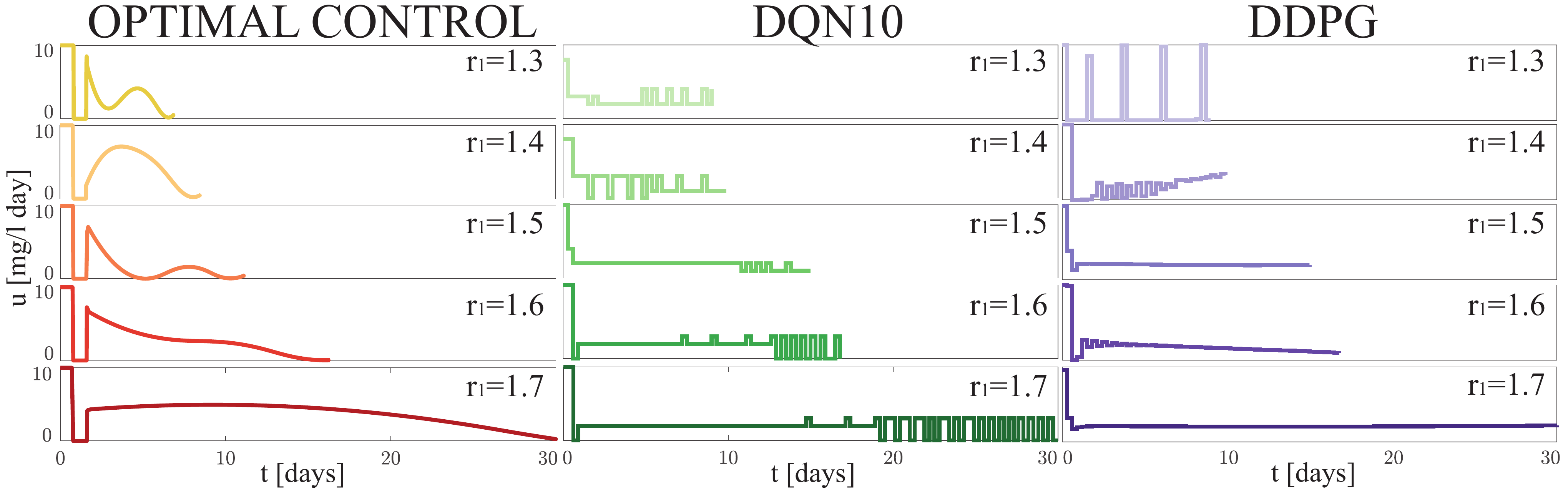}
	\caption{Optimal policies provided by the different methods when sweeping the values of $r_1$.}
	\label{fig:CasePatient_sweepR1}
\end{figure*} 
\subsection{Case Patient}
\begin{figure}[h!]
	\centering
	\includegraphics[width=\columnwidth]{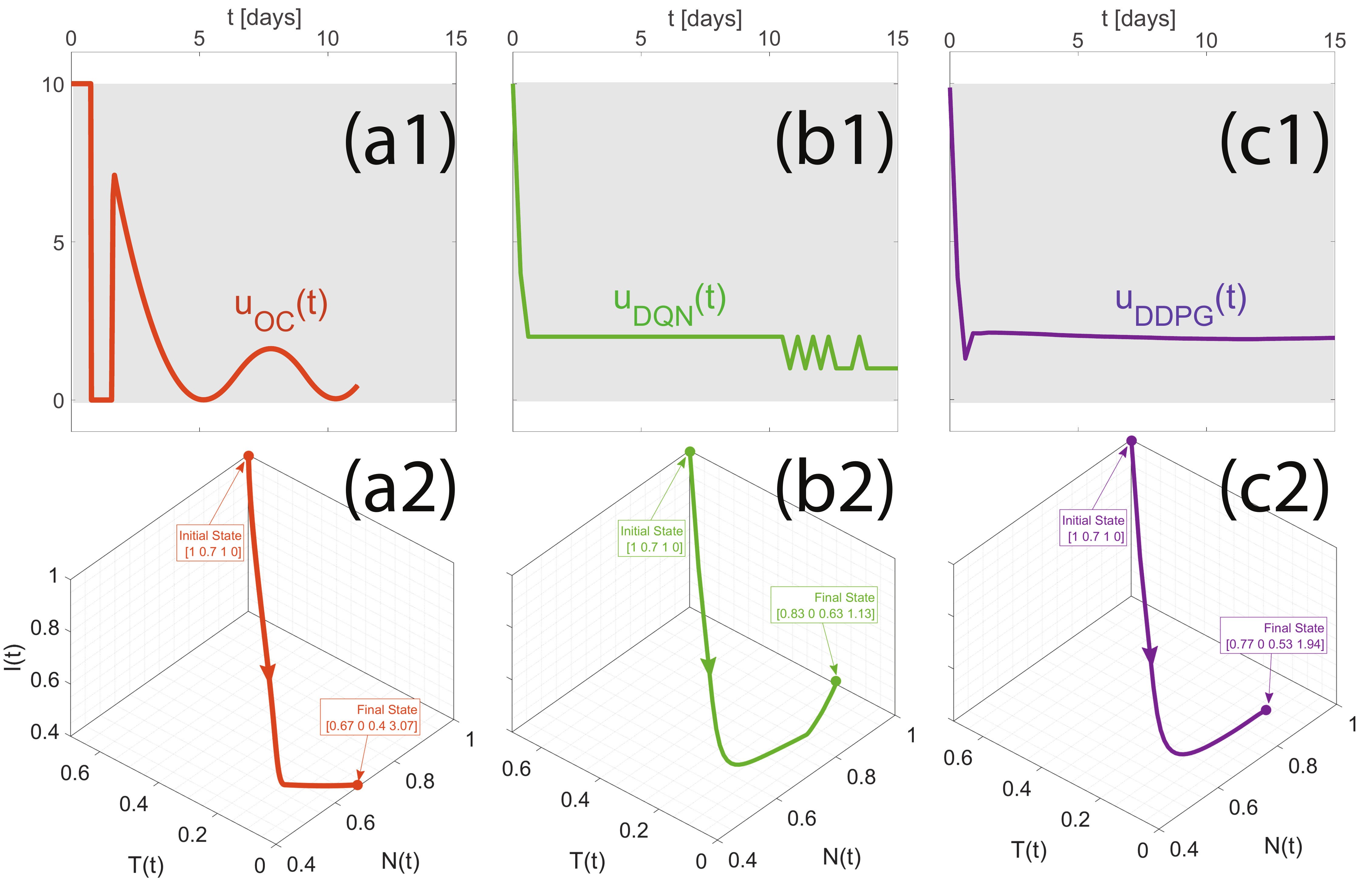}
	\caption{Results for the \textit{Case Patient}. (a) O.C. (b) DQN (c) DDPG. Row 1 provides the policies and row 2 the state evolution when the corresponding policy is applied. The shaded regions represent the feasible control region. }
	\label{fig:CasePatient_results}
\end{figure}

In this case, we add the constraints $N(t)\geq0.4$ and $I(t)\geq0.4$ to the program. The optimal policies found by the three algorithms for the nominal growth tumor rate ($r_1 = 1.5 $) are shown in Fig.  \ref{fig:CasePatient_results}. All the policies are able to exterminate the cancer in $\sim 15$ days. Note also that the policies $u(t)$ provided by the different algorithms share some resemblances: Maximum allowed drug infusion at the beginning of the treatment, followed by an average value $\sim2.5 \text{mg} \cdot l^{-1} \cdot \text{day}^{-1}$. In this case, the cost of the solutions provided by DDPG and DQN is slightly higher than that of O.C.

\begin{figure}[h!]
	\centering
	\includegraphics[width=\columnwidth]{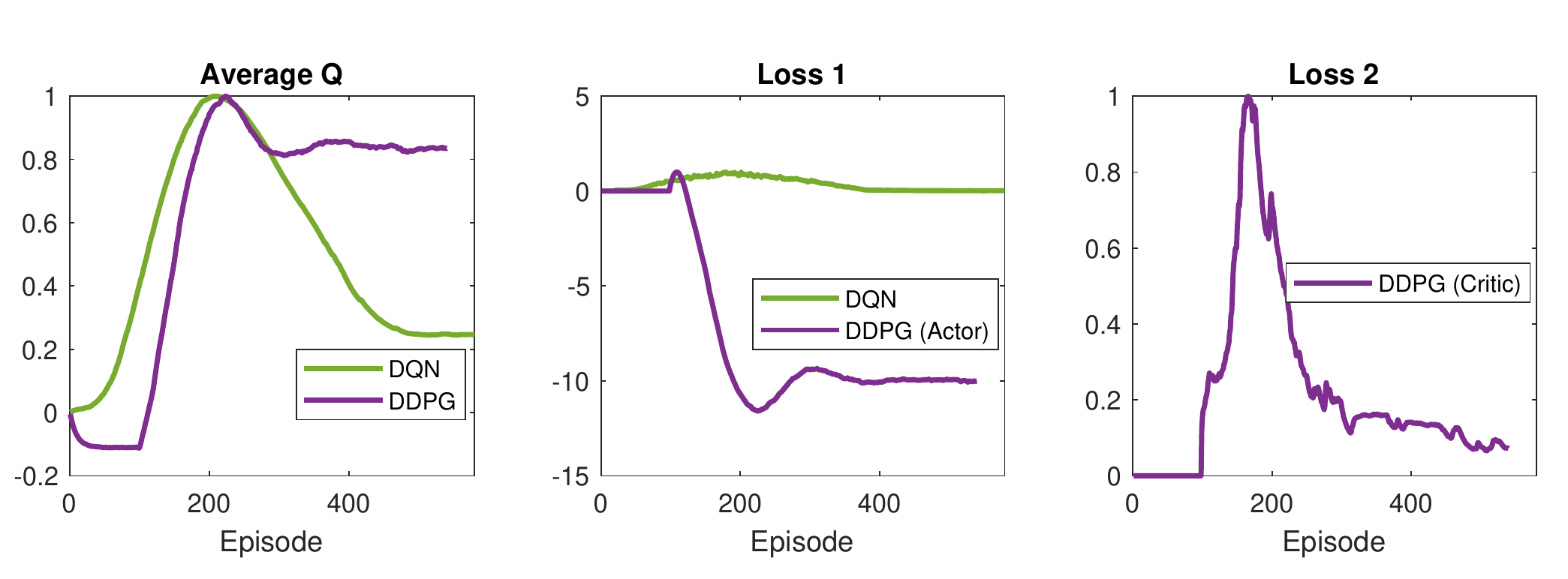}
	\caption{Convergence rates in both DQN10 and DDPG. For DDPG, both the Critic and Actor losses are shown. The values of these plots are normalized.}
	\label{fig:Convergence}
\end{figure}

The convergence rates for DDPG and DQN are shown in Fig. \ref{fig:Convergence}. The training was stopped when the average of the Q-value reached a stationary behaviour. The training for each case considered took $\sim20$ minutes. We also note that the shape of the average Q (see Figure \ref{fig:Convergence}) is that expected from a successful training procedure (see \cite{tampuu2017multiagent} for example): After the replay buffer has been filled, the average Q increases at the beginning of the training, it reaches a maximum (where the agent overestimates the Q value), and then decreases to achieve a stationary value. \\

\subsubsection{Parametric uncertainty: Model Parameter}

To evaluate the sensitivity of the optimal policy to changes in $r_1$, we perturbed $r_1$ around its nominal value and obtained the range for which an optimal policy exists. The problem is feasible for values $r_1 \le1.7$ and the corresponding policies are plotted in Fig. \ref{fig:CasePatient_sweepR1}.  Again, DDPG and DQN obtain a similar policy to that provided by O.C., although some of them present an oscillatory behaviour. Both for DQN and DDPG, these oscillations may be due to the length of the time step ($0.3$ days), and might be reduced by refining it. Moreover, for the case of DQN the coarse discretization (10 nodes) of the action space should be accounted for.\\

\begin{figure}[h!]
	\centering
	\includegraphics[width=\columnwidth]{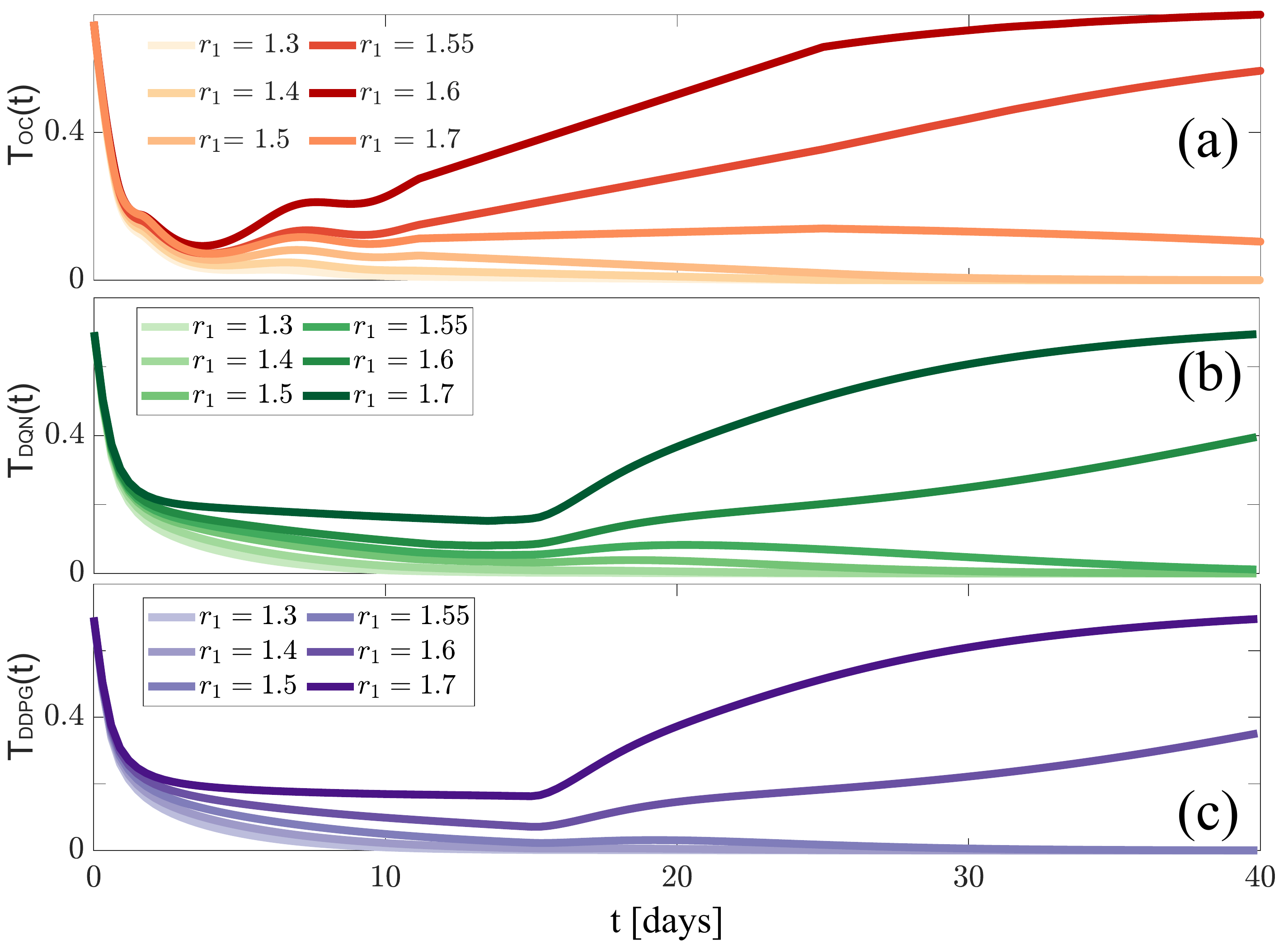}
	\caption{Test of the performance of the nominal policies (computed/trained with $r_1=1.5$) to perturbations in the value of $r_1$. (a) O.C. (b) DQN10 (c) DDPG. DRL methods show improved robustness to perturbations than O.C.}
	\label{fig:Test_r1}
\end{figure}

Once we obtained the values of $r_1$ for which the problem is feasible, we tested the optimal policy obtained for $r_1=1.5$ in different scenarios, where $ r_1\neq 1.5$. The results are shown in Fig. \ref{fig:Test_r1}. In all the algorithms, the policy found eradicates the cancer for $r_1\le1.5$, and does not cure it for $r_1\ge1.6$. However, for the case $r_1=1.55$, DRL algorithms are successful exterminating the cancer, but O.C. is not.  The  increased robustness of DRL to parametric uncertainty might be due to its \textit{exploration} phase: exploration makes the policies found by DRL algorithms \textit{contain} more information than the optimal control solution, which only focuses on finding a minimum for the given model. \\

\subsubsection{Parametric uncertainty: initial condition}
\begin{figure}[h!]
	\centering
	\includegraphics[width=\columnwidth]{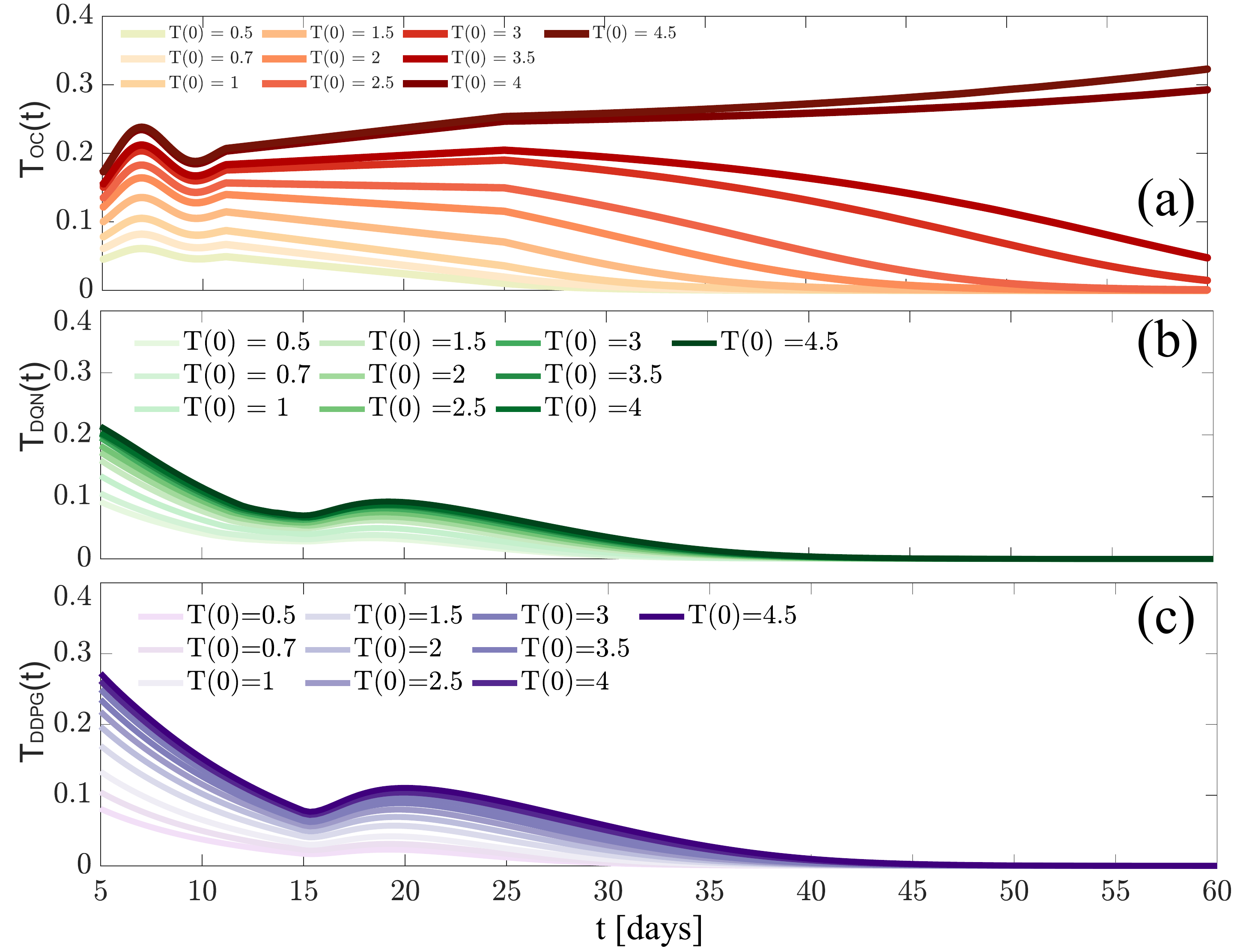}
	\caption{Evolution of the tumor cell population when the policy obtained for the nominal initial condition (computed/trained with $T_0 = 0.7$) is applied for perturbed initial size of the tumor. (a) O.C. (b) DQN10 (c) DDPG. It is remarkable that while optimal control does not manage to exterminate the disease in all the cases, DRL methods do, showing increased robustness. The first 5 days have been removed from the graph for the sake of visualization.}
	\label{fig:IC_sweepTest}
\end{figure}

For this case, the agent is trained with the nominal initial condition $T_0=0.7$, and  tested with perturbed values of $T_0$. The perturbed values and corresponding results are shown in Fig. \ref{fig:IC_sweepTest}. Note that DRL is able to exterminate the cancer for all the cases, while O.C. fails to do it for $T_0\ge3.5$. Again, this argues for  the increased robustness of the policies provided by DRL in comparison to O.C. \\

\subsubsection{Stochastic forcing in tumor dynamics }
\begin{figure}[h!]
	\centering
	\includegraphics[width=0.7\columnwidth]{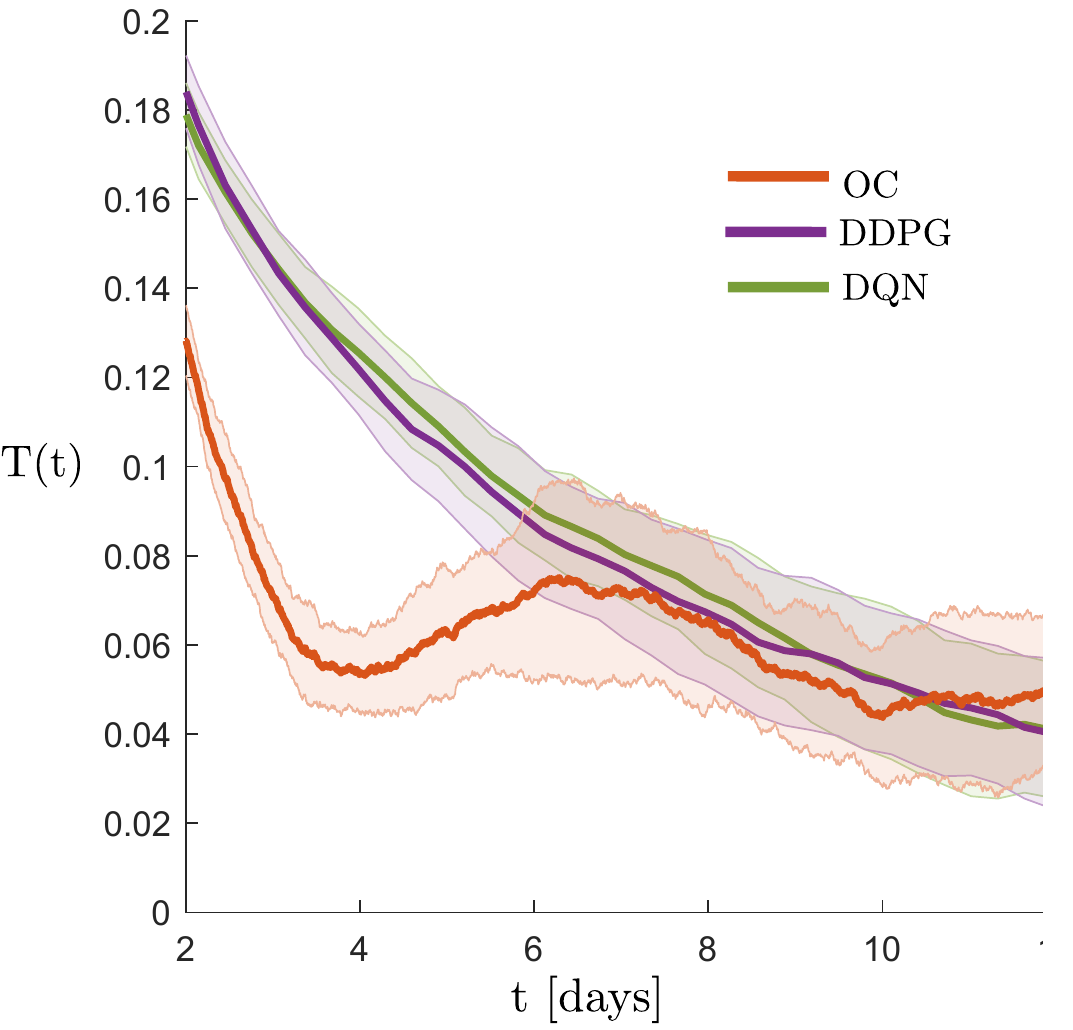}
	\caption{Evolution of $T(t)$ for the noisy case. Note the improved performance of DDPG and DQN at $t=t_f$.}
	\label{fig:stochasticity_merged}
\end{figure}

In this last scenario, the agent is tested with a stochastic forcing term in the dynamics of $T(t)$. The plots of the mean and standard deviations of the solutions found are shown in Fig. \ref{fig:stochasticity_merged}. DQN and DDPG drive the tumor closer to extermination than O.C. at the end of the treatment. \\ 

\begin{figure}[]
	\centering
	\includegraphics[width=.9\columnwidth]{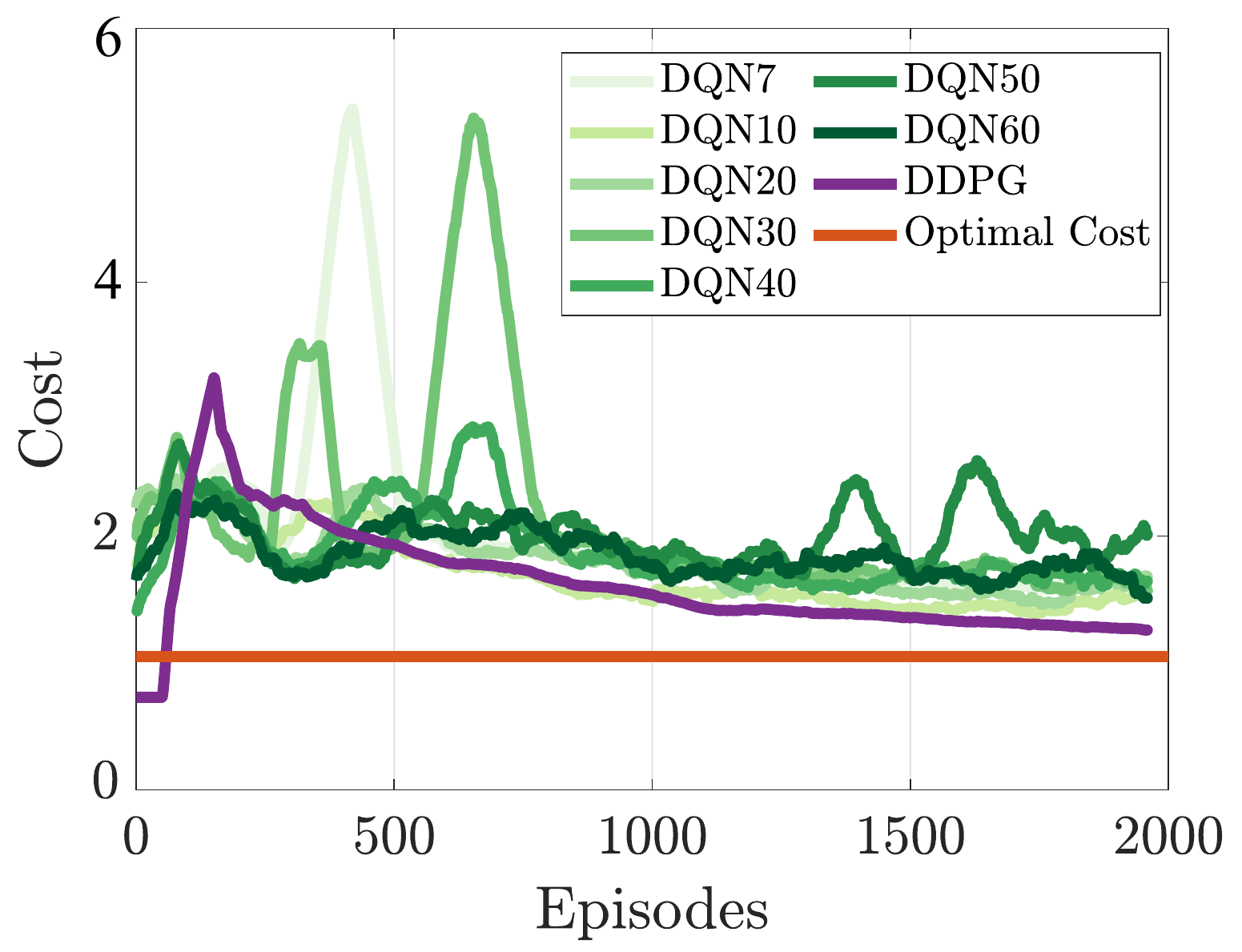}
	\caption{Mean cost (-reward) of the policies found as a function of the training episodes for each algorithm compared to that of the optimal controller. Legend indicates the algorithms with the respective training episodes.}
	\label{fig:sampling}
\end{figure} 

\subsection{Sampling}
An interesting question to investigate is the dependence of learning in DQN and DDPG on the number of episodes. This question is specially relevant in data-poor scenarios, where one may wonder if the size of the dataset is enough to obtain a suitable policy. With this aim, the agent was evaluated after each training episode, and its cost was compared with  that of O.C. In the case of DQN, seven different discretizations of the action space  were considered: 7, 10, 20, 30, 40, 50 and 60 nodes. The resulting plot is shown in Fig. \ref{fig:sampling}. It is observed that the DRL costs asymptote to the optimal one and that, after $\sim 1500$ episodes, DDPG obtains a cost that is close to that of Optimal Control. Note also that, in general, DDPG tends to obtain better agents than DQN.

%\FloatBarrier
\section{CONCLUSIONS AND FUTURE WORK}
\label{sec:conclusions_future_work}
This work presented a comparison between classical O.C. and model-free DRL approaches, both in discrete and continuous action space. We showed that, with an accurate model of the dynamics, O.C. provides the best solution, but closely followed by DRL. Moreover, we showed that in the presence of parametric or stochastic uncertainty, DRL approaches have the potential to outperform classical trajectory optimization O.C. techniques.  \\

%% Unconstrained and constrained cases
In the \textit{Case 0}, all the algorithms found the same optimal policy. In the \textit{Case Patient}, the policies found by DRL perform similarly to O.C, but they exhibit increased robustness to uncertainties. Regarding the relative merits of DQN and DDPG, we found that DDPG outperforms DQN, presumably due to its continuous action space. Furthermore, it seems to learn faster. \\

%%% Sampling
The sampling analysis of the algorithms showed that approximately 1500 calls to the model are needed for DDPG to obtain a performance close to optimal. \\

As future work, we consider comparing the perfomance of DDPG and DQN with robust optimal control, where uncertainty on the dynamics is taken into account in the formulation of the optimal control program at the expense of increasing the cost associated to the policy found. Moreover, we plan also to compare the performace of DDPG and DQN with Model Predictive Control, that benefits from feedback in each iteration.

\iffalse
\textbf{Future work:} Given the encouraging results of this work, we consider it would be advisable to train and test the DRL algorithms with real patient data. If they also perform well in that scenario, \textit{in-vivo} clinical DRL would be recommended.
\fi


%%%%%%%%%%%%%%%%%%%%%%%%%%%%%%%%%%%%%%%%%%%%%%%%%%%%%%%%%%%%%%%%%%%%%%%%%%%%%%%%

%%%%%%%%%%%%%%%%%%%%%%%%%%%%%%%%%%%%%%%%%%%%%%%%%%%%%%%%%%%%%%%%%%%%%%%%%%%%%%%%

%%%%%%%%%%%%%%%%%%%%%%%%%%%%%%%%%%%%%%%%%%%%%%%%%%%%%%%%%%%%%%%%%%%%%%%%%%%%%%%%

\section*{ACKNOWLEDGMENT}

Both authors would like to thank Prof. David Sontag, Alejandro Rodriguez-Ramos and Dong-Ki Kim for valuable discussions and ideas. They are also grateful for the economic support of Fundacion Bancaria ``la Caixa".

%\newpage

%%%%%%%%%%%%%%%%%%%%%%%%%%%%%%%%%%%%%%%%%%%%%%%%%%%%%%%%%%%%%%%%%%%%%%%%%%%%%%%%
%\balance

\makeatletter
\def\endthebibliography{%
	\def\@noitemerr{\@latex@warning{Empty `thebibliography' environment}}%
	\endlist
}
\makeatother

\bibliographystyle{unsrt}
\bibliography{ref}

\end{document}